\documentclass[10pt,twocolumn,letterpaper]{article}

\usepackage{cvpr}
\usepackage{times}
\usepackage{epsfig}
\usepackage{graphicx}
\usepackage{amsmath}
\usepackage{amssymb}
\usepackage{makecell}

\usepackage[pagebackref=true,breaklinks=true,letterpaper=true,colorlinks,bookmarks=false]{hyperref}

\cvprfinalcopy 

\def\cvprPaperID{****} 

\ifcvprfinal\fi
\begin{document}

\title{EmotionNet Nano: An Efficient Deep Convolutional Neural Network Design for Real-time Facial Expression Recognition}

\author{James Ren Hou Lee, Linda Wang, and Alexander Wong\\
Department of Systems Design Engineering, University of Waterloo, Canada\\
Waterloo Artificial Intelligence Institute, Canada\\
DarwinAI Corp., Canada\\
{\tt\small \{jrhlee, linda.wang, a28wong\}@uwaterloo.ca}
}

\maketitle

\begin{abstract}

While recent advances in deep learning have led to significant improvements in facial expression classification (FEC), a major challenge that remains a bottleneck for the widespread deployment of such systems is their high architectural and computational complexities.  This is especially challenging given the operational requirements of various FEC applications, such as safety, marketing, learning, and assistive living, where real-time requirements on low-cost embedded devices is desired. Motivated by this need for a compact, low latency, yet accurate system capable of performing FEC in real-time on low-cost embedded devices, this study proposes EmotionNet Nano, an efficient deep convolutional neural network created through a human-machine collaborative design strategy, where human experience is combined with machine meticulousness and speed in order to craft a deep neural network design catered towards real-time embedded usage.  Two different variants of EmotionNet Nano are presented, each with a different trade-off between architectural and computational complexity and accuracy.  Experimental results using the CK+ facial expression benchmark dataset demonstrate that the proposed EmotionNet Nano networks demonstrated accuracies comparable to state-of-the-art in FEC networks, while requiring significantly fewer parameters (e.g., 23$\times$ fewer compared to \cite{wang2019facial} at a higher accuracy).  Furthermore, we demonstrate that the proposed EmotionNet Nano networks achieved real-time inference speeds (e.g. $>25$ FPS and $>70$ FPS at 15W and 30W, respectively) and high energy efficiency (e.g. $>1.7$ images/sec/watt at 15W) on an ARM embedded processor, thus further illustrating the efficacy of EmotionNet Nano for deployment on embedded devices.

\end{abstract}

\section{Introduction}

\begin{figure}[t]\center
  \includegraphics[width=0.47\textwidth]{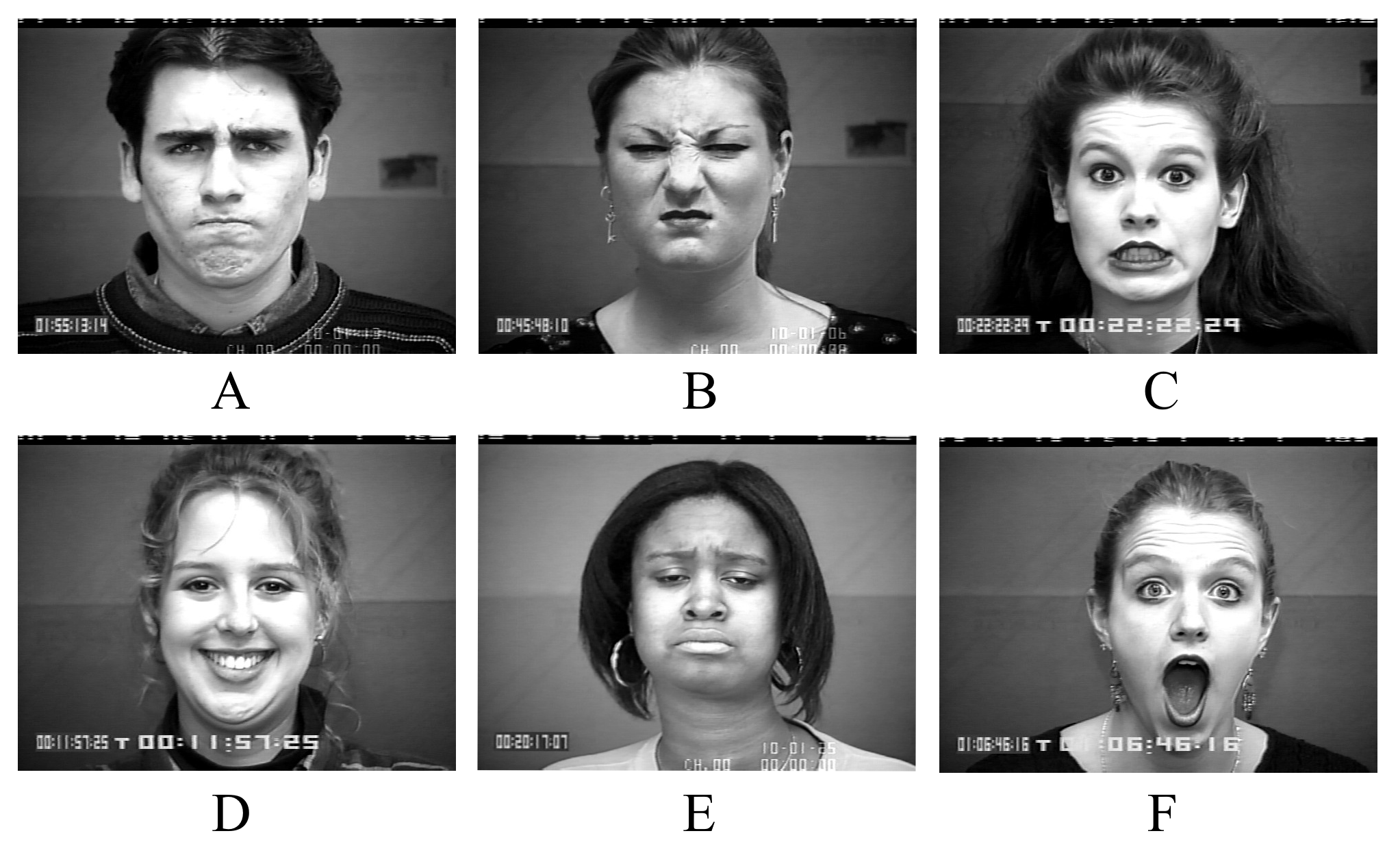}
  \caption{Examples of six different expressions, taken from the CK+ dataset, that needs to be distinguished via facial expression classification: A) anger, B) disgust, C) fear, D) happiness, E) sadness, and F) surprise.}
  \label{fig:6_basic}
\end{figure}

Facial expression classification (FEC) is an area in computer vision that has benefited significantly from the rapid advances in machine learning, which has enabled data collections comprising a diversity of facial expressions captured of different individuals to be leveraged to learn classifiers for differentiating between different facial expression types. In particular, deep learning when applied to FEC has led to significant improvements in accuracy under complex conditions, such as varying lighting, angle, or occlusion. 

Even though the performance of deep learning-based FEC systems continue to rise, widespread deployment of such systems is limited, with one of the biggest hurdles being the high architectural and computational complexities of the deep neural networks that drive such systems.  This hurdle is particularly limiting for real-time embedded scenarios, where low latency operation is required on the low-cost embedded devices. For example, in the area of assistive technologies for improving quality of life, the majority of individuals using such technologies are unwilling to carry large, bulky, and expensive devices with them during their daily lives, as that would be a big hindrance that limits their ability to leverage the technologies in a seamless manner. As such, the assistive devices must leverage small, low-cost, embedded processors, yet provide low latency to enable real-time feedback to the user. Another example is in-car driver monitoring \cite{jeong2018driver}, where a FEC system would record the driver and determine their current mental state, and warn them if their awareness level is deteriorating. In cases such as these, the difference of a few milliseconds of processing is paramount for the safety of not only the user, but also other drivers on the road. In applications for fields such as marketing or security, real-time processing is important to provide salespeople or security guards immediate feedback such that an appropriate response can be made as soon as possible. For those relying on software assistance for social purposes, information is required at no delay in order to keep a conversation alive and not cause discomfort for both parties. 

Motivated by the desire to design deep neural network architectures catered for real-time embedded facial expression recognition, in this study we explore the efficacy of leveraging a human-machine collaborative design strategy that leverages human experience and ingenuity with the raw speed and meticulousness of machine driven design exploration, in order to find the optimal balance between accuracy and architectural and computational complexity.  The resulting deep neural network architecture, which we call \textbf{EmotionNet Nano}, is specifically tailored for real-time embedded facial expression recognition and created via a two phase design strategy. The first phase focuses on leveraging residual architecture design principles to capture the complex nuances of facial expressions, while the second phase employed machine-driven design exploration to generate the final tailor-made architecture design that achieves high architectural and computational efficiency while maintaining a high performance. We present two variants of EmotionNet Nano, each with a different trade-off between accuracy and complexity, and evaluate both variants on the CK+ \cite{CK+} benchmark dataset against state-of-the-art facial expression classification networks.

The paper is organized as follows.  Section 2 discusses related work in the area of facial expression classification and efficient deep neural network architecture design. Section 3 presents in detail the methodology leveraged to design the proposed EmotionNet Nano.  Section 4 presents in detail the network architecture of EmotionNet Nano and explores interesting characteristics observed in the overall design.  Section 5 presents the experiments conducted to evaluate the efficacy of EmotionNet Nano in terms of accuracy, architectural complexity, speed, and energy efficiency.  Section 6 provides a discussion on not only performance but also social implications of EmotionNet Nano.  Finally, conclusions are drawn and future directions are discussed in Section 7.

\section{Related Work}
A variety of deep neural network architectures have been proposed for FEC, ranging from deep convolutional neural networks (DCNN) to recurrent neural networks (RNN) \cite{CNN-RNN} to long-short term memory (LSTM) \cite{lstm} and have been explored, but those introduced in literature have generally required significant architectural complexity and computational power in order to detect and interpret the nuances of human facial expressions. As an alternative to deep learning, strategies leveraging other machine learning strategies such as Support Vector Machines (SVM) \cite{michel-svm} and hand-crafted features such as Local Binary Patterns (LBP) \cite{happy2012real, shan2005recognizing}, dense optical flow \cite{bargal2016emotion}, Histogram of Oriented Gradients (HOG) \cite{kumar2016real}, or Facial Action Coding System \cite{Ekman1978FacialAC} have also been explored in literature, but generally have been shown to achieve lower accuracy when compared to deep learning-based approaches, which can better learn the subtle differences that exist between human facial expressions.

To mitigate the aforementioned hurdle and improve widespread adoption of powerful deep learning-driven approaches for FEC in real-world applications, a key direction that is worth exploring is the design of highly efficient deep neural network architectures tailored for the task of real-time embedded facial expression recognition.  A number of strategies for designing highly efficient architectures have been explored.  One strategy is reducing the depth of the neural network architecture~\cite{khorrami2015deep} to reduce computational and architectural complexity; more specifically, neural networks with a depth of just five were leveraged to learn discriminating facial features.  Another strategy is reducing the input resolution of the neural network architecture, with Shan et. al. \cite{shan2005recognizing} showing that FEC can be performed even at low image resolutions of 14 x 19 pixels, which can further reduce the number of operations required for inference by a large margin. Despite the improved architectural or computational efficiencies gained by leveraging such efficient network design strategies, they typically lead to noticeable reductions in facial expression classification accuracy and as such alternative strategies that enable a better balance between accuracy, architectural complexity, and computational complexity are highly desired.

More recently, there has been a focus on human-driven design principles for efficient deep neural network architecture design, ranging from depth-wise separable convolutions \cite{chollet2017xception} to Inception \cite{szegedy2015going} macroarchitectures to residual connections \cite{resnet}.  Such design principles can substantially improve FEC performance while reducing architectural complexity~\cite{pramerdorfer2016facial}. However, despite the improvements gained in architectural efficiency, one challenge with human-driven design principles is that it is quite time consuming and challenging for humans to hand-craft efficient neural network architectures that are tailored for specific applications such as FEC that possesses a strong balance between a high performance accuracy, fast inference speed, and low memory footprint, primarily due to the sheer complexity of neural network behaviours under different architectural configurations.

In an attempt to address this challenge, neural architecture search (NAS) strategies have been introduced to automate the model architecture engineering process by finding the maximally performing network design from all possible network designs within a search space. However, given the infinitely large search space within which the optimal network architecture may exist in, significant human effort is often required in designing the search space in a way that reduces it to a feasible size, as well as defining a search strategy that can run within desired operational constraints and requirements in a reasonable amount of time. Therefore, a way to combine both human-driven design principles and machine-driven design exploration is highly desired and can lead to efficient architecture designs catered specifically to FEC.

\section{Methods}
In this study, we present EmotionNet Nano, a highly efficient deep convolutional neural network architecture design for the task of real-time facial emotion classification for embedded scenarios. EmotionNet Nano was designed using a human-machine collaborative strategy in order to leverage both human experience as well as the meticulousness of machines. The human-machine collaborative design strategy leveraged to create the proposed EmotionNet Nano network architecture design is comprised of two main design stages: i) principled network design prototyping, and ii) machine-driven design exploration.

\subsection{Principled Network Design Prototyping}

In the first design stage, an initial network design prototype, $\varphi$, was designed using human-driven design principles in order to guide the subsequent machine-driven exploration design stage. In this study, the initial network design prototype of EmotionNet Nano leveraged residual architecture design principles \cite{resnet}, as it was previously demonstrated to achieve strong performance on a variety of recognition tasks. More specifically, the presence of residual connections within a deep neural network architecture have been shown to provide a good solution to both the vanishing gradient and curse of dimensionality problems. Residual connections also enable networks to learn faster and easier, with little additional cost to architectural or computational complexity. Additionally, as the network architecture depth increases, each consecutive layer should perform no worse than its previous layer due to the identity mapping option. As a result, residual network architecture designs have been shown to work well for the problem of FEC \cite{hasani2017facial, zhou2019facial,khorrami2015deep}.  In this study, the final aspects of the initial network design prototype, $\varphi$, consists of an average pooling operation followed by a fully connected softmax activation layer to produce the final expression classification results.  The final macroarchitecture and microarchitecture designs of the individual modules and convolutional layers of the proposed EmotionNet Nano were left to the machine-driven design exploration stage to design in an automatic manner. To ensure a compact and efficient real-time model catered towards embedded devices, this second stage was guided by human-specified design requirements and constraints targeting embedded devices possessing limited computational and memory capabilities. 

\subsection{Machine Driven Design Exploration}

Following the initial human-driven network design prototyping stage, a machine-driven design exploration stage was employed to determine the macroarchitecture and microarchitecture designs at the individual module level to produce the final EmotionNet Nano. In order to determine the optimal network architecture based on a set of human defined constraints, generative synthesis \cite{wong2018ferminets} was leveraged for the purpose of machine-driven design exploration. Defined in Equation \ref{eq:gensynth}, we can formulate generative synthesis as a constrained optimization problem, where the goal is to find a generator $\mathcal{G}$ that, given a set of seeds $\mathcal{S}$, can generate networks $\{\mathcal{N}_s | s \in \mathcal{S}\}$ that maximize a universal performance function $\mathcal{U}$ while also satisfying constraints defined in an indicator function $1_r(\cdot)$,

\begin{equation}
    \mathcal{G} = \max_{\mathcal{G}} \mathcal{U}(\mathcal{G}(s)) \text{ subject to } 1_r(\mathcal{G}(s)) = 1, \forall s \in \mathcal{S}
    \label{eq:gensynth}
\end{equation}

\noindent As such, given a human-defined indicator function $1_r(\cdot)$ and an initial network design prototype $\varphi$, generative synthesis is guided towards learning generative machines that generate networks within the human-specified constraints. 

An important factor in leveraging generative synthesis for machine-driven design exploration is to define the operational constraints and requirements based on the desired task and scenario in a quantitative manner via the indicator function $1_r(\cdot)$.  In this study, in order to learn a compact yet highly efficient facial expression classification network architecture, the indicator function  $1_r(\cdot)$ was set up such that: i) accuracy $\geq$ 92\% on CK+~\cite{CK+}, and ii) network architecture complexity $\leq$ 1M parameters. These constraint values were chosen to explore how compact a network architecture for facial expression classification can be while still maintaining sufficient classification accuracy for use in real-time embedded scenarios.  As such, we use the accuracy of Feng \& Ren \cite{feng2018dynamic} as the reference baseline for determining the accuracy constraint in the indicator function. 

\section{EmotionNet Nano Architecture}
\begin{figure*}
  \includegraphics[width=\textwidth]{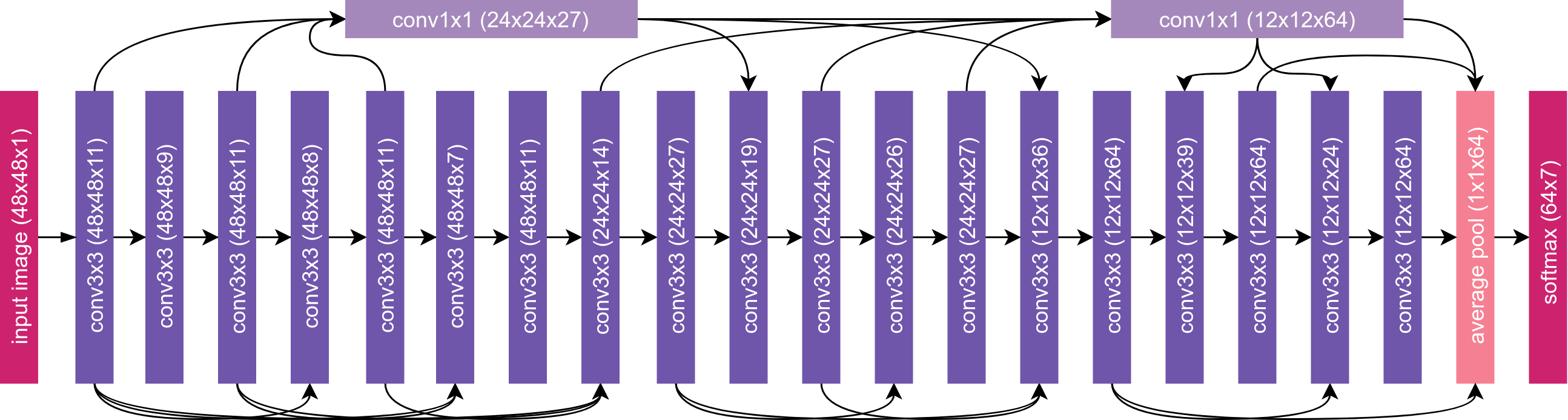}
  \caption{\textbf{EmotionNet Nano Architecture.} The network architecture exhibits high macroarchitecture and microarchitecture heterogeneity, customized towards capturing deep facial features.  Furthermore, the network architecture exhibits selective long-range connectivity throughout the network architecture.  The number of channels per layer are based on EmotionNet Nano-B.}
  \label{fig:arch}
\end{figure*}

The network architecture of the proposed EmotionNet Nano is shown in Figure \ref{fig:arch}. A number of notable characteristics of the proposed EmotionNet Nano network architecture design are worth discussing as they give insights into architectural mechanisms that strike a strong balance between complexity and accuracy.  

\subsection{Architectural heterogeneity} 
A notable characteristic about the architecture that allows the network to achieve high efficiency even with a low number of parameters is the macroarchitecture and microarchitecture heterogeneity. Unlike hand-crafted architecture designs, the macroarchitecture and microarchitecture designs within the EmotionNet Nano network architecture as generated via machine-driven design exploration differ greatly from layer to layer. For instance, there are a mix of convolution layers with varying shapes and different number of channels per layer depending on the needs of the network. As shown in Figure \ref{fig:arch}, there are a greater number of channels needed as the sizes of feature maps decrease. 

The benefit of high microarchitecture and macroarchitecture heterogeneity in the EmotionNet Nano network architecture is that it enables different parts of the network architecture to be tailored to achieve a very strong balance between architectural and computational complexity while maintaining model expressiveness in capturing necessary features. The architectural diversity in EmotionNet Nano demonstrates the advantage of leveraging a human-collaborative design strategy as it would be difficult for a human designer, or other design exploration methods to customize a network architecture to the same level of architectural granularity.

\subsection{Selective long-range connectivity}

Another notable characteristic of the EmotionNet Nano network architecture is that it exhibits selective long range connectivity throughout the network architecture. The use of long range connectivity in a very selective manner enables a strong balance between model expressiveness and ease of training, and computational complexity.  Most interesting and notable is the presence of two densely connected $1\times1$ convolution layers that take in outputs from multiple $3\times3$ convolution layers as input, with its output connected farther down at later layers.  Such a $1\times1$ convolution layer design provides dimensionality reduction while retaining salient features of the channels through channel mixing, thus further improving architectural and computational efficiency while maintaining strong model expressiveness.

\section{Experimental Results}
To evaluate the efficacy of the proposed EmotionNet Nano, we examine the network complexity, computational cost and classification accuracy against other facial expression classification networks on the CK+ \cite{CK+} dataset, which is the most extensively used laboratory-controlled FEC benchmark dataset \cite{li2018deep, mmi}.

\subsection{Dataset}

The Extended Cohn-Kanade (CK+) \cite{CK+} dataset contains 593 video sequences from a total of 123 different subjects, ranging from 18 to 50 years of age with a variety of genders and heritage. Each video shows a facial shift from the neutral expression to a targeted peak expression, recorded at 30 frames per second (FPS) with a resolution of either 640x490 or 640x480 pixels. Out of these videos, 327 are labelled with one of seven expression classes, anger, contempt, disgust, fear, happiness, sadness, and surprise. The CK+ database is widely regarded as the most extensively used laboratory-controlled FEC database available, and is used in the majority of facial expression classification methods \cite{li2018deep, mmi}.  Figure~\ref{fig:6_basic} shows that the CK+ dataset has good diversity for each expression type, which is important from an evaluation perspective.  However, as the CK+ dataset does not provide specific training, validation, and test set splits, a mixture of splitting techniques can be observed in literature. For experimental consistency, we adopt the most common dataset creation strategy where the last three frames of each sequence is extracted and labeled with the video label~\cite{li2018deep}.  In this study, we performed subject-independent 10-fold cross validation on the resulting 981 facial expression images.

\begin{figure}[h]
  \includegraphics[width=0.45\textwidth]{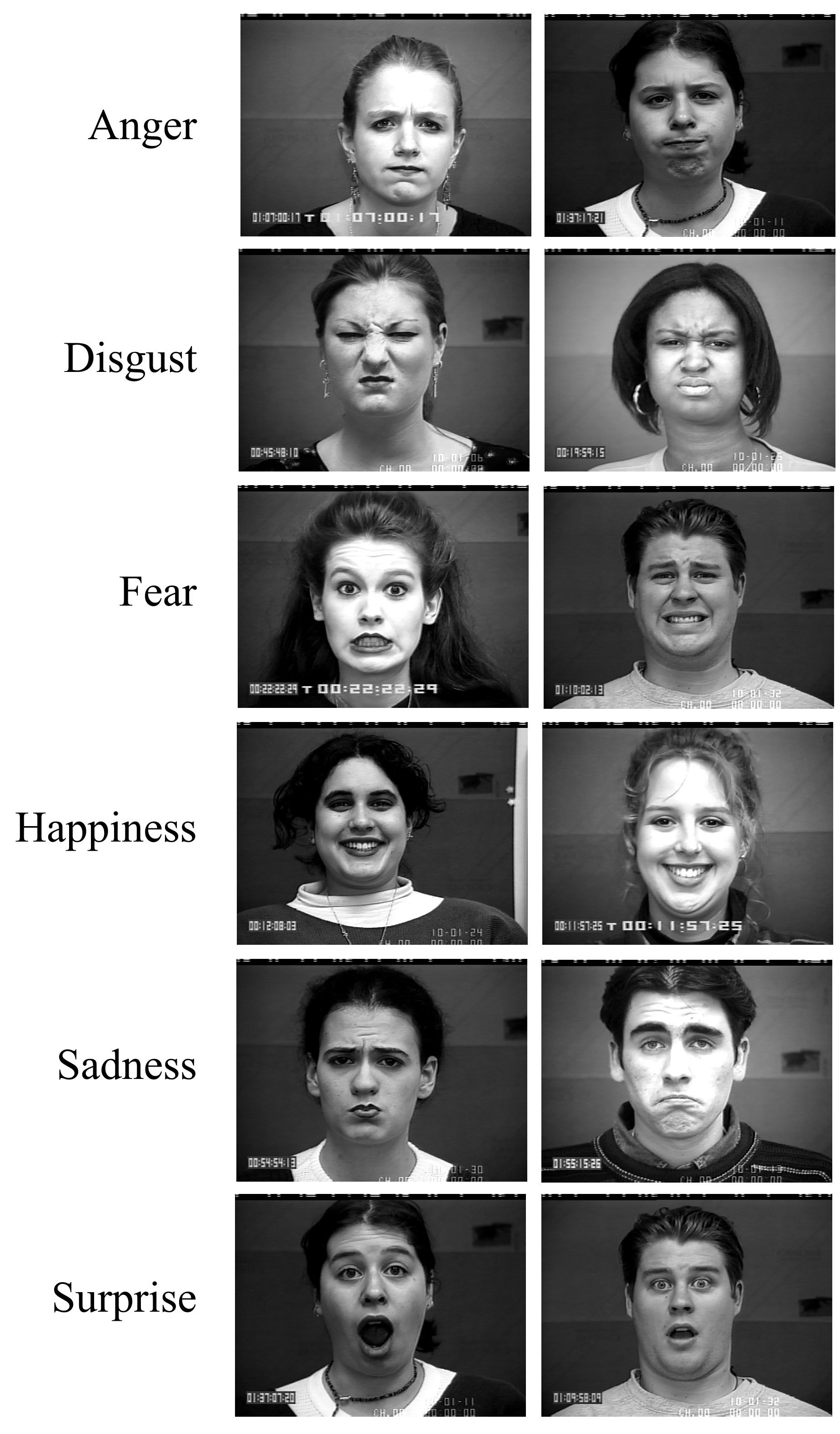}
  \caption{\textbf{Diversity of expressions in the CK+ dataset.} Example faces for each expression type in CK+ is shown. Contempt not included as relevant subjects did not give publication consent. }
  \label{fig:6_basic}
\end{figure}

\subsection{Implementation Details}

EmotionNet Nano was trained for 200 epochs using an initial learning rate of $1\mathrm{e}{-3}$, multiplied by $1\mathrm{e}{-1}$, $1\mathrm{e}{-2}$, $1\mathrm{e}{-3}$, and $0.5\mathrm{e}{-3}$ at epochs 81, 121, 161, and 181 respectively. Categorical cross-entropy loss was used with the Adam \cite{kingma2014adam} optimizer. Data augmentation was applied to the inputs, including rotation, width and height shifts, zoom, and horizontal flips. Following this initial training, we leveraged a machine-driven exploration stage to fine tune the network specifically for the task of FEC. Training was performed using a GeForce RTX 2080 Ti GPU. The Keras \cite{chollet2015keras} library was leveraged for this study.

\subsection{Performance Evaluation}

Two variants of EmotionNet Nano were created to examine the different trade-offs between architectural and computational complexity and accuracy. In order to demonstrate the efficacy of the proposed models in a quantitative manner, we compare the performance of both variants against state-of-the-art facial expression classification networks introduced in literature, shown in Table \ref{tab:static_comp_table}.  It can be observed that both EmotionNet Nano-A and Nano-B networks achieve strong classification accuracy, with EmotionNet Nano-A in particular achieving comparable accuracy with the highest-performing state-of-the-art networks that are more than a magnitude larger.  While EmotionNet Nano-B has lower accuracy than the highest-performing networks, it is still able to achieve comparable accuracy as \cite{feng2018dynamic} while being three orders of magnitude smaller.  A more detailed discussion of the performance comparison will be provided in the next section; overall, it can be observed that both EmotionNet Nano variants provide the greatest balance between accuracy and complexity, making it well-suited for embedded scenarios. 

\begin{table*}[ht]
\caption{\textbf{Comparison of facial expression classification networks on the CK+ dataset.}  We report 10-fold cross-validation average accuracy on the CK+ dataset with 7 classes (anger, contempt, disgust, fear, happiness, sadness, and surprise).}
\begin{center}
\begin{tabular}{|c|c|c|}
\hline
\textbf{Method} & \textbf{Params (M)} & \textbf{Accuracy (\%)}\\
\Xhline{2\arrayrulewidth}
Ouellet \cite{ouellet2014real} & 58 & 94.4 \\
\hline
Feng \& Ren \cite{feng2018dynamic} & 332 & 92.3 \\
\hline
Wang \& Gong \cite{wang2019facial} & 5.4 & 97.2 \\
\hline
Otberdout et al. \cite{otberdout2019automatic} & 11 & 98.4 \\
\hline
\hline
EmotionNet Nano-A & 0.232 & 97.6 \\
\hline
EmotionNet Nano-B & 0.136 & 92.7 \\
\hline
\end{tabular}\par
\bigskip
\label{tab:static_comp_table}
\end{center}
\vspace{-0.3in}
\end{table*}

\begin{figure}[t]
  \includegraphics[width=0.5\textwidth]{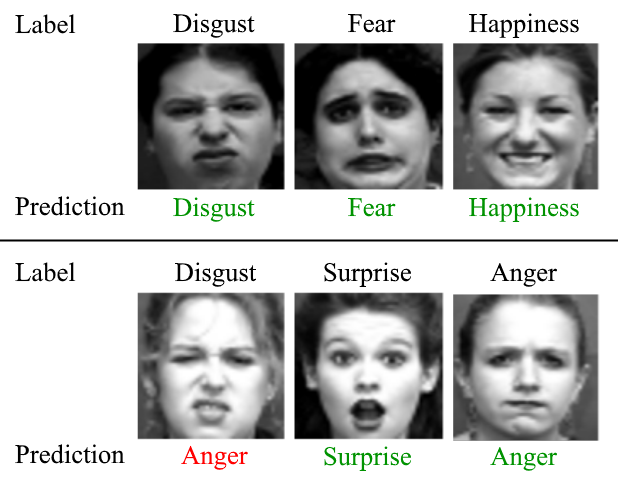}
  \caption{\textbf{Example expression predictions of faces in the CK+ dataset using EmotionNet Nano-A.} Five of the faces are classified correctly, indicated in green, with an example of a misclassified expression (disgust), shown in red.}
  \label{fig:example_classifications}
\end{figure}

The distribution of expressions in CK+ is unequal, which results in an unbalanced dataset both for training and testing. The effects of this are prevalent when classifying the contempt or fear expressions, both of which are underrepresented in CK+ (e.g. there are only 18 examples of contempt, whereas there are 83 examples of surprise). Due to the nature of human facial expressions, similarities between expressions do exist, but the networks are generally able to learn the high-level distinguishing features that separate one expression from another. However, incorrect classifications can still occur, as shown in Figure \ref{fig:example_classifications}, where a ``disgust'' expression is falsely predicted to be ``anger.''

\subsection{Speed and Energy Efficiency}
We also perform a speed and energy efficiency analysis, shown in Table \ref{tab:energy_eff}, to demonstrate the efficacy of EmotionNet Nano in real-time embedded scenarios. Here, an ARM v8.2 64-Bit RISC embedded processor was used for evaluation. Referring to Table \ref{tab:energy_eff}, both EmotionNet Nano variants are able to perform inference at $>$25 FPS and $>$70 FPS on the tested embedded processor at 15W and 30W respectively, which more than fulfills a real-time system constraint.  In terms of energy efficiency, both EmotionNet Nano variants demonstrated high power efficiency, with the Nano-B variant running at 5.29 images/sec/watt on the embedded processor.  

\begin{table}[h]
\caption{\textbf{EmotionNet Nano Speed and Energy Efficiency.} All metrics are computed on an ARM v8.2 64-Bit RISC embedded processor at different power levels.}
\begin{center}
\begin{tabular}{|c|c|c|c|c|}
\hline
 & \multicolumn{2}{c}{\textbf{15W}} & \multicolumn{2}{|c|}{\textbf{30W}}\\
\hline&&&&\\[-0.85em]
\textbf{Model} & \textbf{FPS} & \textbf{{[}$\frac{\text{images/s}}{\text{watt}}${]}} & \textbf{FPS} & \textbf{{[}$\frac{\text{images/s}}{\text{watt}}${]}}\\[3pt]
\Xhline{2\arrayrulewidth}
EmotionNet Nano-A & 25.8 & 1.72 & 70.1 & 2.34\\
\hline
EmotionNet Nano-B & \textbf{32.8} & \textbf{2.19} & \textbf{72.9} & \textbf{2.43}\\
\hline
\end{tabular}\par
\bigskip
\label{tab:energy_eff}
\end{center}
\vspace{-0.3in}
\end{table}

\section{Discussion}
In this study, we explore the human-machine collaborative design of a deep convolutional neural network architecture capable of performing facial expression classification in real-time on embedded devices.  It is important to note that other extremely fast deep convolutional neural network architectures exist, such as MicroExpNet~\cite{ccuugu2017microexpnet}, which is capable of processing 1851 FPS on an Intel i7 CPU, is less than 1 MB in size, and is tested on the CK+ 8 class problem (7 facial expression classes plus neutral) on which it achieves 84.8\% accuracy. Although a motivating result, a direct comparison cannot be made with EmotionNet Nano as well as other facial expression classification networks evaluated in this study due to the different class sizes. 

Compared against state-of-the-art facial expression classification network architectures tested on CK+ using the same seven expression classes (see Figure \ref{tab:static_comp_table}), both variants of the proposed EmotionNet Nano are at least an order of magnitude smaller yet provide comparable accuracy to state-of-the-art network architectures. For example, EmotionNet Nano-A is $>$23$\times$ smaller than \cite{wang2019facial}, yet achieves higher accuracy by 0.4\%.  Furthermore, while EmotionNet Nano-A achieves an accuracy that is 0.8\% lower than the top-performing \cite{otberdout2019automatic}, it possesses $>$47$\times$ fewer parameters.  In the case of EmotionNet Nano-B, it achieved higher accuracy (by 0.4\%) than \cite{feng2018dynamic} while having three orders of magnitude fewer parameters.  

Looking at the experimental results around inference speed and energy efficiency on an embedded processor at different power levels (see Table \ref{tab:energy_eff}), it can be observed that both variants of EmotionNet Nano achieved real-time performance  and high energy efficiencies.  For example, in the case of EmotionNet Nano-A, it was able to exceed 25 FPS and 70 FPS at 15W and 30W, respectively, with energy efficiencies exceeding 1.7 images/s/watt and 2.34 images/s/watt at 15W and 30W, respectively.  This demonstrates that the proposed EmotionNet Nano is well-suited for high-performance facial expression classification in real-time embedded scenarios. An interesting observation that is worth noting is the fact that while the inference speed improvements of EmotionNet Nano-B over EmotionNet Nano-A exceeds 27\% at 15W, there is only a speed improvement of 4\% at 30W.  As such, it can be seen that EmotionNet Nano-B is more suitable at low-power scenarios but at high-power scenarios the use of EmotionNet Nano-A is more appropriate given the significantly higher accuracy achieved.

\subsection{Implications and Concerns}

The existence of an efficient facial expression classification network of running in real-time on embedded devices can have an enormous impact in many fields, including safety, marketing, and assistive technologies. In terms of safety, driver monitoring or improved surveillance systems are both areas that benefit from higher computational efficiency, as it lowers the latency between event notifications as well as reduces the probability that a signal will be missed. With a real-time facial expression classification system in the marketing domain, companies will gain access to enhanced real-time feedback when demonstrating or promoting a product, either in front of live audiences or even in a storefront. The largest impact however, is likely in the assistive technology sector, due to the increased accessibility that this efficiency provides. The majority of individuals do not have access to powerful computing devices, nor are they likely to be willing to carry a large and expensive system with them as it would be considered an inconvenience to daily living. 

\begin{figure}[h]\center
  \includegraphics[width=0.48\textwidth]{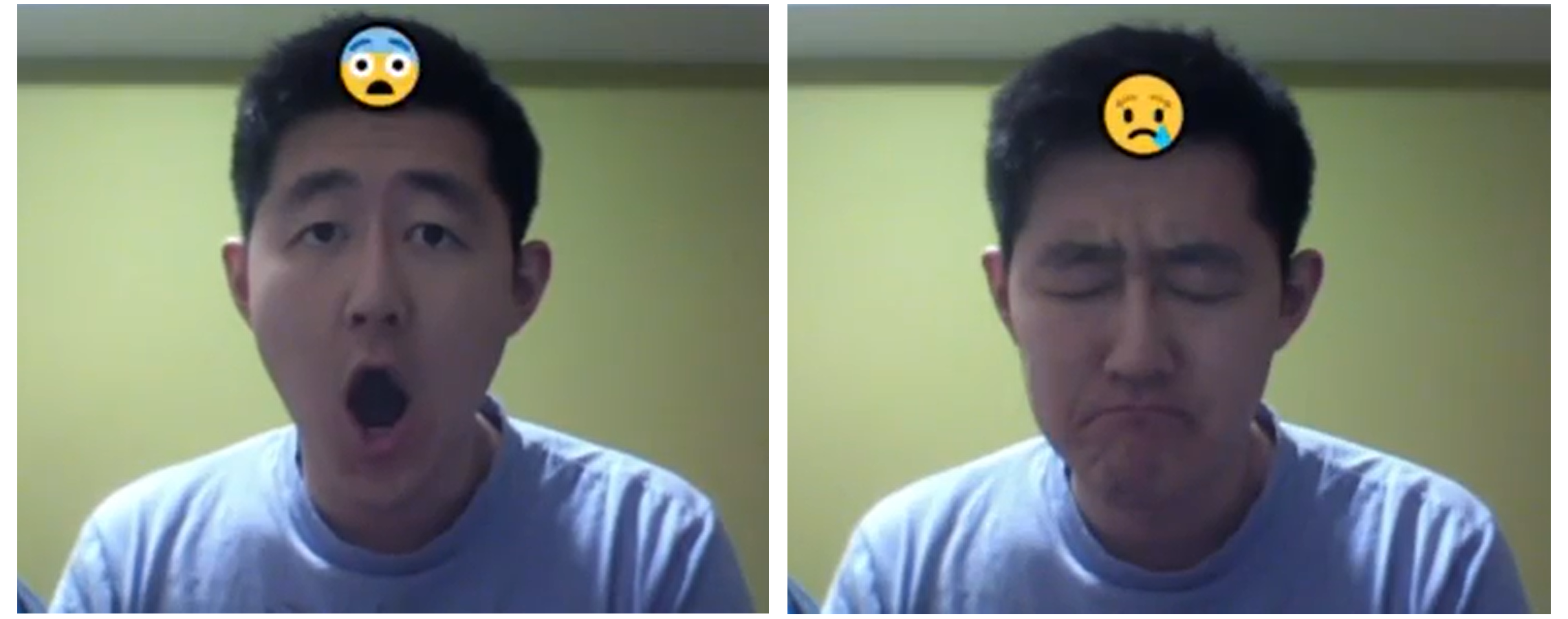}
  \caption{\textbf{Assistive technology for Autistic Spectrum Disorder.} Example of how EmotionNet Nano can be leveraged to assist individuals with Autistic Spectrum Disorder to better infer emotional states from facial expressions during social interactions in the form of augmented reality. }
  \label{fig:assistive}
\end{figure}

As shown in this study, EmotionNet Nano can achieve accurate real-time performance on embedded devices at a low power budget, granting the user access to a facial expression classification system on their smartphone or similar edge device with embedded processors without rapid depletion of their battery. This can be extremely beneficial towards tasks such as depression detection, empathetic tutoring, or ambient interfaces, and can also help individuals who suffer from Autistic Spectrum Disorder better infer emotional states from facial expressions during social interaction in the form of augmented reality (see Figure \ref{fig:assistive} for a visual illustration of how EmotionNet Nano can be used to aid in conveying emotional state via an augmented reality overlay).

Although EmotionNet Nano has many positive implications, there exist concerns that must be considered before deployment. The first concern is privacy, as individuals may dislike being on camera, even if no data storage is taking place. Privacy concerns, especially ones centered around filming without consent, are likely to arise if these systems start to be used in public areas. The combination of facial expression classification together with facial recognition could result in unwanted targeted advertising, even though this could be seen as a positive outcome for some. Additionally, wrong classifications could result in unintended implications. When assisting a user in an ambient interface or expression interpretation task, a misclassified expression could result in a negative experience with major consequences. For example, predicting ``sad'' or ``angry'' expressions as ``happy'' could influence the user to behave in the wrong manner.  These concerns and issues are all worth further exploration and investigation to ensure that such systems are used in a responsible manner.

\section{Conclusion}
In this study, we introduced EmotionNet Nano, a highly efficient deep convolutional neural network design tailored for facial expression classification in real-time embedded scenarios by leveraging a human-machine collaborative design strategy.  By leveraging a combination of human-driven design principles and machine-driven design exploration, the EmotionNet Nano architecture design possesses several interesting characteristics (e.g., architecture heterogeneity and selective long-range connectivity) that makes it tailored for real-time embedded usage.  Two variants of the proposed EmotionNet Nano network architecture design were presented, both of which achieve a strong balance between architecture complexity and accuracy while illustrating performance trade-offs at that scale.  Using the CK+ dataset, we show that the proposed EmotionNet Nano can achieve comparable accuracy to state-of-the-art facial expression classification networks (at 97.6\%) while possessing a significantly more efficient architecture design (possessing just 232K parameters).  Furthermore, we demonstrated that EmotionNet Nano can achieve real-time inference speed on an embedded processor at different power levels, thus further illustrating its suitability for real-time embedded scenarios.

Future work involves incorporating temporal information into the proposed EmotionNet Nano design when classifying video sequences. Facial expressions are highly dynamic and transient in nature \cite{3dc}, meaning that information about the previous expression is valuable when predicting the current expression. Therefore, the retention of temporal information can lead to increased performance, at the expense of computational complexity.  Investigating this trade-off between computational complexity and improved performance when leveraging temporal information would be worthwhile. 

{\small
\bibliographystyle{ieee_fullname}
\bibliography{egbib}
}

\end{document}